\documentclass[11pt,a4paper]{article}
\usepackage[hyperref]{acl2021}
\usepackage{graphicx}
\usepackage{times}
\usepackage{latexsym}

\usepackage{amsmath}
\usepackage{microtype}
\usepackage{multirow}
\aclfinalcopy 

\title{AdaST: Dynamically Adapting Encoder States in the Decoder for End-to-End Speech-to-Text Translation}

\author{Wuwei Huang, Dexin Wang, Deyi Xiong  \Thanks{~Corresponding author}\\
  College of Intelligence and Computing, Tianjin University, Tianjin, China\\
  \texttt{ \{wwhuang, 2019218007, dyxiong\}@tju.edu.cn} \\
}
\date{}

\begin{document}
\maketitle
\begin{abstract}

In end-to-end speech translation, acoustic representations learned by the encoder are usually fixed and static, from the perspective of the decoder, which is not desirable for dealing with the cross-modal and cross-lingual challenge in speech translation. In this paper, we show the benefits of varying acoustic states according to decoder hidden states and propose an adaptive speech-to-text translation model that is able to dynamically adapt acoustic states in the decoder. We concatenate the acoustic state and target word embedding sequence and feed the concatenated sequence into subsequent blocks in the decoder. In order to model the deep interaction between acoustic states and target hidden states, a speech-text mixed attention sublayer is introduced to replace the conventional cross-attention network. Experiment results on two widely-used datasets show that the proposed method significantly outperforms state-of-the-art neural speech translation models.
\end{abstract}

\begin{figure*}[ht]
\centering
\includegraphics[width=.7\textwidth,height=5.6cm]{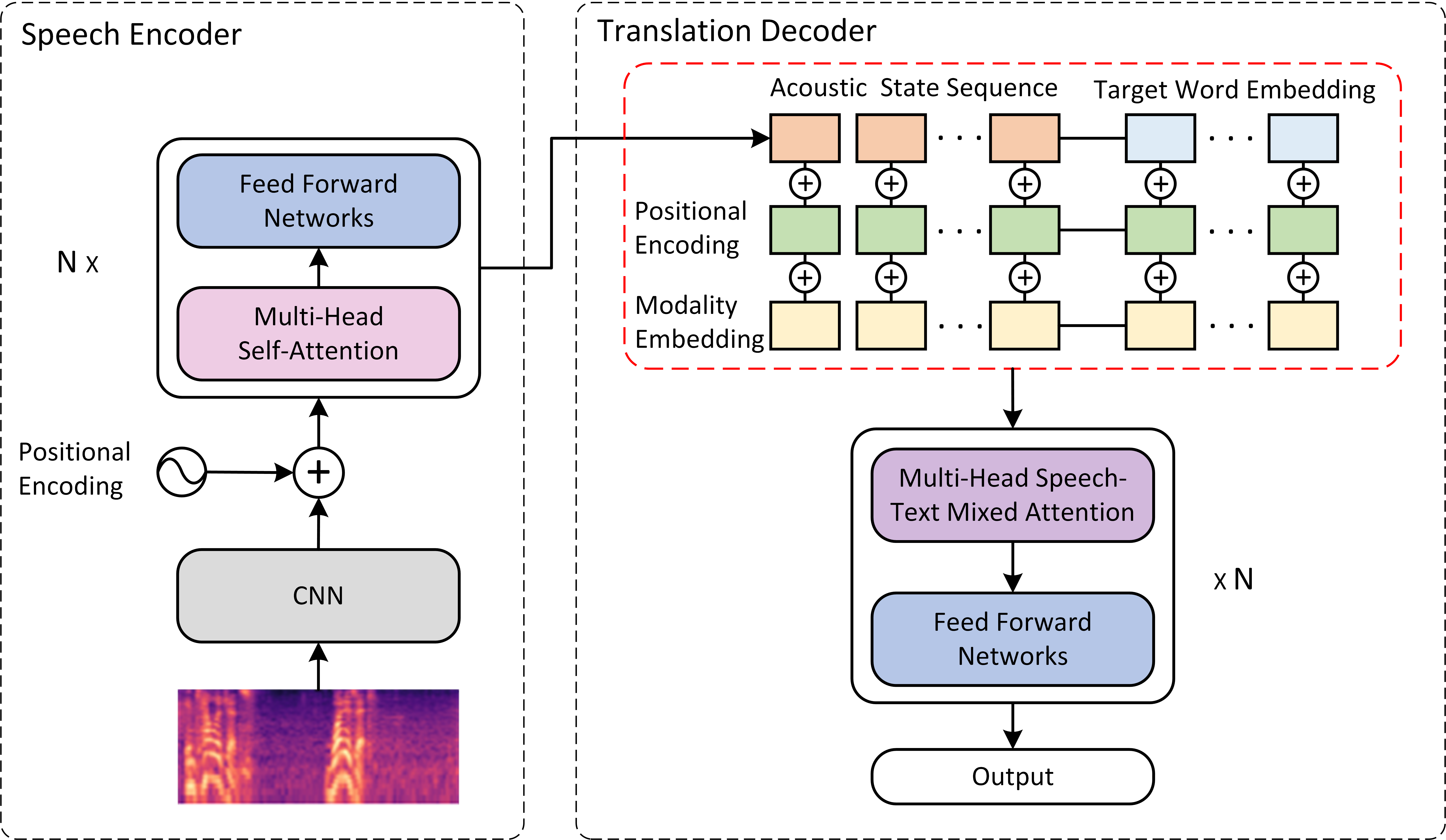}
\caption{Diagram of the proposed AdaST that dynamically adapts acoustic encoder states in the decoder for end-to-end ST.}\label{fig:basemodel}
\end{figure*}

\section{Introduction}
Speech-to-text translation (ST) aims at translating the source language speech into the text of the target language. Approaches to ST can be roughly divided into two categories: end-to-end ST and cascaded ST. Early research on ST is primarily using a cascaded model that combines a speech recognition (ASR) module with a machine translation component, both usually trained independently on speech and parallel corpora \citep{DBLP:conf/icassp/Ney99,DBLP:conf/interspeech/MatusovKN05}. In contrast, end-to-end ST, which directly translates the speech of the source language into text of the target language \citep{DBLP:journals/corr/BerardPSB16}, not only avoids error propagation in the ASR-MT pipeline, but also greatly reduces inference latency.

 However, despite these advantages, end-to-end ST is confronted with its own challenging problems:  performing cross-modal translation and cross-lingual conversion in one shot. On the one hand, compared with text-to-text translation, end-to-end ST has to deal with acoustic inputs which  are typically longer than their corresponding text inputs. This makes the cross-modal source-target dependencies more complicated. On the other hand, compared with monotonic ASR, end-to-end speech translation usually handles non-monotonic cross-lingual conversion.

 Generally, end-to-end ST uses the seq2seq encoder-decoder framework \citep{DBLP:conf/nips/SutskeverVL14}  as the backbone for training and inference, where the encoder computes hidden states layer by layer according to speech inputs. The decoder yields target translations word by word by attending to the fixed-after-computing hidden states of the encoder. Since the hidden states are static in the encoder, information only flows one direction: from the encoder to the decoder. Given the cross-modal and cross-lingual challenge in end-to-end ST, we argue that more sophisticated interaction between the encoder and decoder would be desirable.

 In this paper, we propose an adaptive ST (AdaST) model that incorporates acoustic states into the decoder for modeling the deep interaction between the encoder and decoder for end-to-end ST. We enable AdaST to dynamically adapt encoder states in the decoder when target hidden states are updated layer by layer. It also learns to represent speech and text in one shared space in the decoder for mitigating the cross-modal issue.

Our contributions can be summarized as follows:
\begin{itemize}
\item We present AdaST, a new architecture for end-to-end ST, which learns representations of two modalities (textual and audio) in one shared space in the decoder.
\item We conduct experiments to validate the effectiveness of AdaST. Our experiments and analyses disclose that dynamically adaptive acoustic representations are more desirable than static acoustic states for end-to-end ST.
\end{itemize}


\section{Related Work}

\citet{DBLP:journals/corr/BerardPSB16} demonstrate the potential of end-to-end neural ST and \citet{DBLP:conf/interspeech/WeissCJWC17} achieve good performance by using an end-to-end neural architecture, which trigger more research interests in end-to-end ST. Both \citet{DBLP:conf/naacl/BansalKLLG19} and \citet{DBLP:conf/icassp/StoianBG20} train a speech recognition model first and then use the encoder of ASR to initialize the encoder of speech translation. \citet{DBLP:conf/icassp/JiaJMWCCALW19} synthesize training data  for end-to-end speech translation from MT and ASR dataset. \citet{DBLP:conf/asru/GangiNT19,DBLP:conf/asru/InagumaDKW19} adapt the idea of multilingual machine translation to speech translation. In addition to these methods, \citet {DBLP:journals/corr/abs-1911-08876} use phoneme-level representations instead of speech frame-level representations as input, greatly reducing the length of acoustic sequences. Knowledge distillation \citep{DBLP:conf/interspeech/LiuXZHWWZ19}, meta-learning \citep{DBLP:journals/corr/abs-1911-04283}, curriculum learning \citep{DBLP:conf/acl/WangWLZY20}, and two-pass decoding \citep{DBLP:conf/icassp/SungLLL19}, have also been studied in end-to-end speech translation.

To solve the cross-modal and cross-lingual challenges of end-to-end speech translation, \citet{DBLP:conf/aaai/WangWLY020} and \citet{DBLP:journals/corr/abs-2009-09704} propose to use submodules to separately analyze cross-modal and cross-lingual problems in end-to-end ST. Each module introduced solves one problem. Unfortunately, they introduce a large number of extra parameters and rely on a large amount of external data to pre-train each submodule. In contrast, we do not introduce any additional submodules and therefore we do not need external data for pretraining.

\section{The AdaST Model}
In this section, we first introduce the widely-used CNN + Transformer structure as the strong  baseline for end-to-end ST. After that we elaborate the proposed AdaST model.

\subsection{Baseline ST Model}
The CNN + Transformer end-to-end ST model consists of a speech encoder and a translation decoder. The basic building unit of Transformer \citep
 {DBLP:conf/nips/VaswaniSPUJGKP17} is the self-attention mechanism, which  can be formulated as follows:
\begin{equation}\label{equ1}
 {\rm Attention}(Q,K,V)={\rm softmax}(\frac{QK^{T}}{\sqrt{d_{k}}})V
\end{equation}

The speech encoder is composed of  N$_{c}$ CNN layers for encoding acoustic signals and N${e}$ Transformer encoder layers stacked over CNN layers.
The translation decoder consists of  N$_{d}$ Transformer decoder layers.

 The CNN module subsamples acoustic features to fit them into the subsequent Transformer encoder layers. The Transformer encoder layers then learn encoder states from the output of the CNN module, which are fixed during decoding. That is to say, the Transformer decoder layers attend to static Transformer encoder hidden states for yielding target words.

\subsection{AdaST}
As shown in Figure \ref{fig:basemodel},  our proposed AdaST uses the same speech encoder as the baseline ST model. The significant difference lies in the decoder. In order to make acoustic states dynamically adaptive to decoder states in each layer,  we  concatenate the hidden acoustic state sequence generated from the last layer of the speech encoder with the target word embedding sequence and feed the concatenated sequence into the subsequent decoder blocks. The concatenated input sequence is combined  with positional encoding, similar to the vanilla Transformer decoder. In addition to positional encoding, we also adapts modality embeddings, which are defined in a embedding matrix with size of $2$ $\times$ $c$ ($c$ is the dimension of attention) adding to the input sequence to distinguish the target textual tokens from the source acoustic features. Modality embeddings has also been used in other cross-modal tasks, e.g., Vilbert for vision-text multimodal pretraining. Our experiments show that using modality embeddings in our model can slightly improve translation quality.

In the decoder, each block consists of a multi-head speech-text mixed attention sublayer and a feedforward sublayer. The multi-head speech-text mixed attention (STMA) is calculated as follows:

\vspace{-2.5ex}
\begin{equation}\label{equ4}
{\rm STMA}(Q,K,V)=
{\rm softmax}(\frac{QK^{T}}{\sqrt{d_{k}}}+ Mask)V
\end{equation}
\vspace{-1ex}
\begin{equation}\label{equ6}
Q={\rm Concat}(src,tgt)W^Q
\end{equation}
\begin{equation}\label{equ7}
K={\rm Concat}(src,tgt)W^K
\end{equation}
\begin{equation}\label{equ8}
V={\rm Concat}(src,tgt)W^V
\end{equation}
where $src$ and $tgt$ represent the sequence of acoustic hidden states and target word embeddings respectively, and $Mask$  is a predefined matrix which serves as  indicators controlling which positions of the acoustic and target sequence are visible to attention heads, similar to the look ahead mask matrix used in Transformer to prevent the decoder from attending future tokens.

In each decoder layer of  the proposed AdaST, we divide the $Mask$ matrix into four parts:

$$Mask =
\begin{bmatrix}
M_{SS} & M_{ST} \\
M_{TS} & M_{TT}
\end{bmatrix}$$

$M_{SS}$ represents the self-attention mask matrix of the acoustic state, which is the same as used in the encoder. $M_{ST}$ is the mask matrix for the attention from acoustic states to target hidden states. During parallel training, as source acoustic states are not visible to target hidden states, we set all values of $M_{ST}$ to minus infinity to forbid such attention. $M_{TS}$ denotes the mask matrix for attention from target hidden states to acoustic states. Values in $M_{TS}$ are the same as the mask matrix used for the cross-attention in Baseline ST. $M_{TT}$ is the mask matrix for self-attention over target hidden states, which is the same as the mask matrix used for self-attention on the Baseline ST decoder.

The proposed AdaST benefits from  the following features. First, the acoustic states and decoder hidden states are unified into a shared semantic space. Second,  the acoustic states at each decoder layer change accordingly after the calculations at the current layer are performed. Third, instead of calculating softmax for self-attention and then calculating softmax for cross-attention in the baseline ST, the neural representations in the AdaST decoder are updated by calculating a single softmax over both acoustic states and hidden states for target words. With these changes, we hope to mitigate the cross-modal and cross-lingual challenges in end-to-end ST.

\section{Experiments}
\label{sec:length}
We conducted experiments  to examine the proposed AdaST model.

\subsection{Datasets}
We used two datasets that are widely adopted to evaluate end-to-end ST: IWSLT18 En-De  and Augmented Librispeech En-Fr \citep{DBLP:conf/icassp/BerardBKP18}.

\textbf{Augmented Librispeech English-French.} The corpus provides triples for each instance:  English speech signal, English transcription, French text translation from the aligned  e-books. Following \citet{DBLP:conf/acl/WangWLZY20}, we only used the 100 hours clean data for training, with 2 hours data as the development set and 4 hours as the test set, which corresponds to 47,271, 1071 and 2048 utterances respectively. To be consistent with their settings, we also doubled the training data by concatenating the aligned references with pseudo translations by the Google Translate.

\textbf{ IWSLT18 English-German.} The IWSLT18 speech translation dataset is from TED Talks \citep{long2020shallowdiscourseannotationchinese,long-etal-2020-ted}, which contains 271 hour speech with 171K corresponding English transcripts and German translations. As there is no validation set in this dataset, we randomly sampled 2000 samples from the training data as our validation set. Following \citet{DBLP:conf/acl/WangWLZY20}, we used tst2013 as the test set.

\subsection{Settings}
We built our model based on the  Espnet toolkit \citep{DBLP:conf/acl/InagumaKDKYHW20}. On the two datasets, we extracted 80-dimensional Fbank features from audio files, setting the step size as 10ms and the window size as 25ms. We deleted sentences with frame size larger than 3000 and sentences with poor alignments. Following  \citet{DBLP:conf/aaai/WangWLY020}, we adopted speed perturbation with factors 0.9 and 1.1. To further reduce overfitting,  we used SpecAugment strategy \citep{DBLP:journals/corr/abs-1911-08876}. In Librispeech, we used subword level decoding, which was performed via SentencePiece with a size of 1K tokens. In IWSLT18, we performed character level decoding. As the tst2013 of IWSLT2018 is not aligned, we employed Espnet¡¯s default LIUM SpkDiarization tool to segment each audio sequence. We used RWTH toolkit \citep{DBLP:conf/iwslt/BenderZMN04} to calculate BLEU scores \citep{DBLP:conf/acl/PapineniRWZ02}.

A two-layer CNN was taken in the speech encoder. The step size was set to 2. The size of the convolution kernel was $2\times2$. The dimension of the attention was set to 256. We used 12-layer encoder. The number of decoder layers in both the baseline and AdaST was set to 10. We used the Adam optimizer \citep{DBLP:journals/corr/KingmaB14} and run our models on four P100 GPUs.

\subsection{Main Results}
In order to make each layer of the decoder to interact with acoustic states, our model requires additional computational overhead. However, the conventional source-to-target attention network in Transformer is subsumed in the decoder, which helps AdaST to use fewer parameters than Transformer, hence partially offsetting the additional cost. Overall, the number of parameters in AdaST is 0.65 million fewer than that of the standard CNN+Transformer structure.  On the augmented dataset, AdaST increased the training time by 11.7\% and the inference time by 15.7\%. We compared our work against previous state-of-the-art models and the ASR pretraining + MT fine-tuning method. Table \ref{tabel:1} shows the results on the two datasets.  We observe that the proposed AdaST is able to achieve improvements of +0.83 BLEU and +1.18 BLEU over the best baseline results on En-Fr and En-De translation, respectively. This demonstrates that our proposed method benefits end-to-end ST  at both the character and subword level. We have also carried out experiments to compare against a standard CNN+Transformer model with deeper encoder and decoder. Experiment results show that simply deepening either encoder or decoder of the standard structure is not helpful for speech-to-text translation.
\begin{table}[t]
\scriptsize

\centering
\begin{tabular}{c|c|c} 

\hline
&Method&BLEU\\
\hline
\multirow{4}*{En-Fr}&LSTM ST \citep{DBLP:conf/icassp/BerardBKP18}& 12.90\\
&Transformer+ASR pre-train \citep{DBLP:conf/acl/InagumaKDKYHW20}&15.53\\
&Transformer+ASR pre-train&16.27\\
&AdaST&\textbf{17.10}\\
\hline
\multirow{4}*{En-De}&Transformer+ASR pre-train \citep{DBLP:conf/acl/InagumaKDKYHW20}&13.12\\
&Transformer+ASR pre-train \citep{DBLP:conf/acl/WangWLZY20}&15.35 \\
&Transformer+ASR pre-train&15.21\\
&AdaST&\textbf{16.39}\\
\hline
\end{tabular}
\caption{Results on the two datasets.}\label{tabel:1}
\end{table}

\begin{table}[t]
\centering

\begin{tabular}{c|c|c} %

\hline
&Structure&Result\\
\hline
\multirow{2}*{ST}&Transformer+ASR pre-train&16.27\\
&AdaST&17.10\\
\hline
\multirow{2}*{ASR}&Transformer&7.5\\
&AdaST&8.3\\
\hline
\multirow{2}*{MT}&Transformer&18.10\\
&AdaST&18.16\\
\hline
\end{tabular}
\caption{Results of using AdaST on different tasks, i.e., speech translation (ST), automatic speech recognition (ASR) and machine translation (MT).  BLEU ($\uparrow$) scores are reported on ST and MT while CER ($\downarrow$) on ASR. }\label{tabel:2}
\end{table}

\begin{table}[t]
\centering
\begin{tabular}{c|c} 

\hline

Structure&BLEU\\
\hline
Transformer+ASR pre-train&15.21\\
Transformer+Additional Self-Att&16.08\\
AdaST&\textbf{16.39}\\

\hline
\end{tabular}
\caption{Results of dynamic vs. static acoustic states.}\label{tabel:3}
\end{table}

\begin{table}[t]
\centering
\begin{tabular}{c|c}
\hline
Method&ACC\\
\hline

Transformer+ASR pre-train&74.2\\
AdaST&96.7\\
\hline
\end{tabular}
\caption{Classification accuracy (\%) on speaker verification.}\label{tabel:4}
\end{table}

\section{Analysis}
We conducted further experiments and analyses to take a deep look into our proposed method.

\subsection{Only Cross-modal or Cross-lingual Challenge}
In order to investigate whether our proposed architecture is helpful for a task with only cross-modal or cross-lingual challenge, we also conducted experiments for automatic speech recognition (ASR)
and machine translation (MT) tasks with the proposed method on the Agmented Librispeech dataset. Experimental results in Table \ref{tabel:2} show that the performance of ASR task drops, while the performance of MT task is improved slightly. This suggests that the proposed architecture is more appropriate for dealing with cross-lingual and cross-modal challenges at the same time.

\subsection{Adaptive vs. Static Acoustic States}
We assume  that dynamically adaptive representations of acoustic states in accord with hidden decoder states at each decoder layer will be of great help to end-to-end ST. In order to examine this hypothesis, we add an additional self-attention at each encoder layer in the baseline ST, which forces acoustic states at the corresponding encoder layer to adapt to decoder hidden states.
 The results on the IWSLT18 dataset, as displayed in Table \ref{tabel:3}, validate this assumption. However, the added additional self-attention  substantially increase the number of parameters at each layer.  By contrast, our AdaST does not introduce additional parameters at each layer to learn adaptive acoustic states on the one hand and achieves better performance on the other.

\subsection{Probing the Speech Encoder}

We further compared the trained speech encoder of our AdaST against that of the baseline ST by evaluating speaker verification accuracy on the Fluent Speech Commands dataset \citep{DBLP:conf/interspeech/LugoschRITB19} to investigate  the change of the semantic information learned by the encoder. Generally, the more semantic information the encoder contains, the less audio information it learns and hence the lower classification accuracy it will obtain. We froze parameters of these two speech encoders, and added a linear classification layer on the top of the encoder . Only the added classification layer is trained on the dataset mentioned above. Table \ref{tabel:4} shows the classification accuracy results, where the baseline encoder achieves 74.2\% while our encoder 96.7\%, substantially higher than the baseline encoder. This indicates that our encoder focuses on modeling the audio modality and passes the major task of modeling semantic information in speech inputs to the decoder. In contrast, the baseline encoder has to model both semantic and modality information of speech inputs, which weakens its modeling capacity for modality and therefore makes it have a much lower performance on speaker verification.

\section{Conclusions}
In this paper, we have presented AdaST, a neural model dynamically adapting acoustic states in the decoder, which is able to mitigate the cross-lingual and cross-modal challenge for end-to-end speech translation.  Experiments demonstrate that AdaST achieves an improvement of 1.18 BLEU points over state-of-the-art neural speech translation models.

\section*{Acknowledgments}
The present research was partially supported by the National Key Research and Development Program of China (Grant No. 2019QY1802).  We would like to thank the anonymous reviewers for their insightful comments.

\bibliographystyle{acl_natbib}
\bibliography{acl2021}


\end{document}